\documentclass[a4paper,fleqn]{cas-dc}

\usepackage[numbers]{natbib}
\usepackage{graphicx}
\usepackage{geometry}
\usepackage{amsmath}

\usepackage{booktabs,makecell}
\usepackage{bbding}
\usepackage{pifont}
\usepackage{wasysym}
\usepackage{amssymb}

\usepackage{url}
\usepackage{xcolor}

\newcommand{\ie}{{\it i.e.}}
\newcommand{\eg}{{\it e.g.}}

\newcommand{\argmin}{\mathop{\rm arg~min}\limits}
\usepackage{hyperref}

\def\tsc#1{\csdef{#1}{\textsc{\lowercase{#1}}\xspace}}
\tsc{WGM}
\tsc{QE}
\tsc{EP}
\tsc{PMS}
\tsc{BEC}
\tsc{DE}

\begin{document}
\let\WriteBookmarks\relax
\def\floatpagepagefraction{1}
\def\textpagefraction{.001}
\shorttitle{ASAD}
\shortauthors{F.Yang et al.}

\title [mode = title]{Actor-identified Spatiotemporal Action Detection --- \\Detecting Who Is Doing What in Videos}                         
\tnotemark[1]


\author[1]{Fan Yang}[orcid=0000-0001-7185-5688]
\cormark[1]
\ead{hongheyangfan@gmail.com}

\author[2]{Norimichi Ukita}

\author[1,3]{Sakriani Sakti}

\author[1,3]{Satoshi Nakamura}

\address[1]{Nara Institute of Science and Technology, Japan}
\address[2]{Toyota Technological Institute, Nagoya, Japan.}
\address[3]{RIKEN, Center for Advanced Intelligence Project AIP, Japan}

\cortext[cor1]{Corresponding author}

\begin{abstract}
The success of deep learning on video Action Recognition (AR) has motivated researchers to progressively promote related tasks from the coarse level to the fine-grained level. Compared with conventional AR that only predicts an action label for the entire video, Temporal Action Detection (TAD) has been investigated for estimating the start and end time for each action in videos. Taking TAD a step further, Spatiotemporal Action Detection (SAD) has been studied for localizing the action both spatially and temporally in videos. However, who performs the action, is generally ignored in SAD, while identifying the actor could also be important. To this end, we propose a novel task, Actor-identified Spatiotemporal Action Detection (ASAD), to bridge the gap between SAD and actor identification.

In ASAD, we not only detect the spatiotemporal boundary for instance-level action but also assign the unique ID to each actor. To approach ASAD, Multiple Object Tracking (MOT) and Action Classification (AC) are two fundamental elements. By using MOT, the spatiotemporal boundary of each actor is obtained and assigned to a unique actor identity. By using AC, the action class is estimated within the corresponding spatiotemporal boundary. Since ASAD is a new task, it poses many new challenges that cannot be addressed by existing methods: i) no dataset is specifically created for ASAD, ii) no evaluation metrics are designed for ASAD, iii) current MOT performance is the bottleneck to obtain satisfactory ASAD results. To address those problems, we contribute to i) annotate a new ASAD dataset, ii) propose ASAD evaluation metrics by considering multi-label actions and actor identification, iii) improve the data association strategies in MOT to boost the MOT performance, which leads to better ASAD results. We believe considering actor identification with spatiotemporal action detection could promote the research on video understanding and beyond. The code is available at \url{https://github.com/fandulu/ASAD}.
\end{abstract}


\maketitle

\section{Introduction}
\label{intro}

Vision-based Action Recognition (AR) aims to detect human-defined actions from a sequence of data (\eg, videos) and has a wide range of applications in our daily life. For instance, it has been applied for YouTube to recognize billions of video tags before recommending a video to us, or for the policemen to quickly retrieval a criminal from thousands-hours surveillance videos, or for a virtual game machine to interact with players, and many others (~\cite{chaquet2013survey,hutchinson2020video}).

\begin{figure}[t!]
	\centering
		\includegraphics[width=\linewidth]{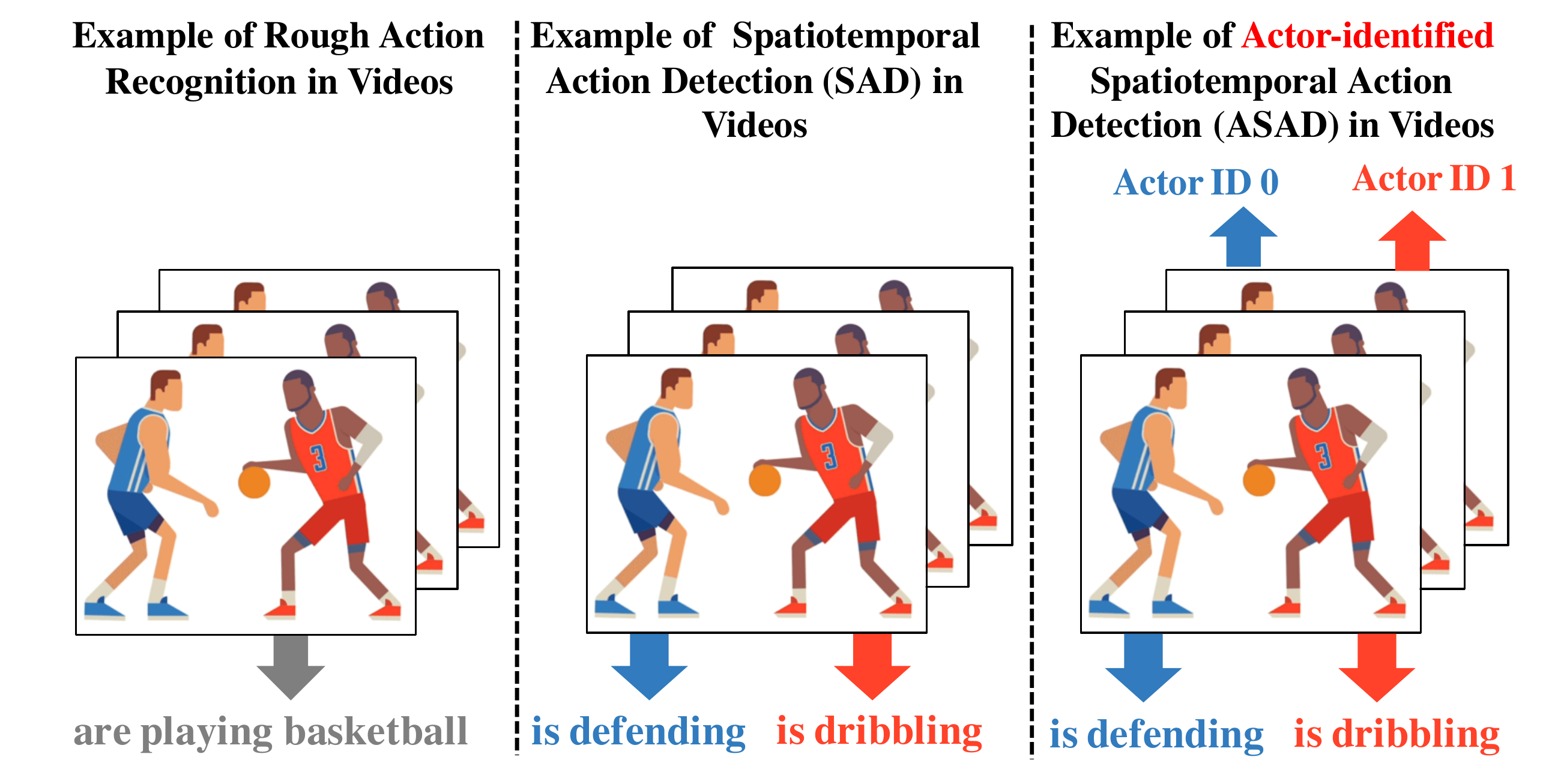}
  \caption{Actor-identified Spatiotemporal Action Detection (ASAD) is Spatiotemporal Action Detection (SAD) pluses actor identification.}
  \label{fig:ASAD_demo}
\end{figure}

In recent years, the success of deep learning on AR has motivated researchers to progressively promote the AR task from the coarse level to the fine-grained level. Compared with conventional AR that only predicts an action label for the entire video, Temporal Action Detection (TAD) has been investigated for estimating the start and end time for each action in videos. Taking TAD a step further, Spatiotemporal Action Detection (SAD) has been studied for localizing the action both spatially and temporally in videos. However, who performs the action is generally ignored in SAD studies. \textbf{We believe actor identification should be considered together with SAD}. When multiple actors are involved in the target scenes (\eg, basketball/soccer games),  it is preferred to know ``who is doing what'', and thus, identifying each actor with their actions is desired. Nonetheless, SAD and actor identification are treated as different tasks for a long time. To this end, we propose a novel task, Actor-identified Spatiotemporal Action Detection (ASAD), to bridge the gap between SAD and actor identification (Figure~\ref{fig:ASAD_demo}).

\begin{figure}[h!]
	\centering
		\includegraphics[width=\linewidth]{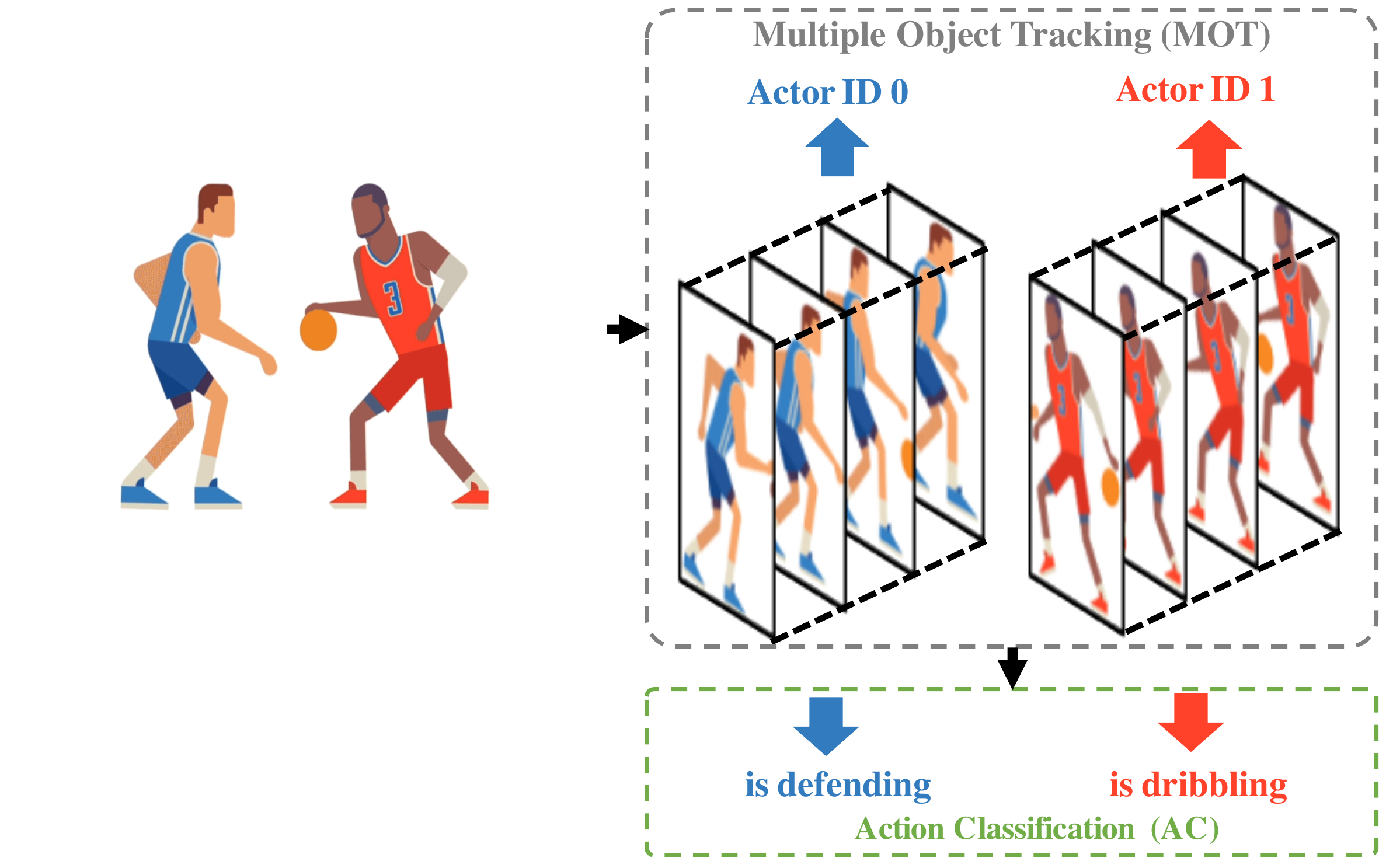}
  \caption{The illustration of ASAD processing.}
  \label{fig:ASAD_pipeline}
\end{figure}

To approach ASAD, Multiple Object Tracking (MOT)~\cite{luo2014multiple} and Action Classification (AC)~\cite{hutchinson2020video} are two fundamental elements (Figure~\ref{fig:ASAD_pipeline}). By using MOT, the spatiotemporal boundary of each actor is obtained and assigned to a unique actor identity. By using AC, the action class is estimated within the corresponding spatiotemporal boundary. In general, they may work as independent modules by considering the model training flexibility.

\begin{figure*}[ht!]
	\centering
		\includegraphics[width=\textwidth]{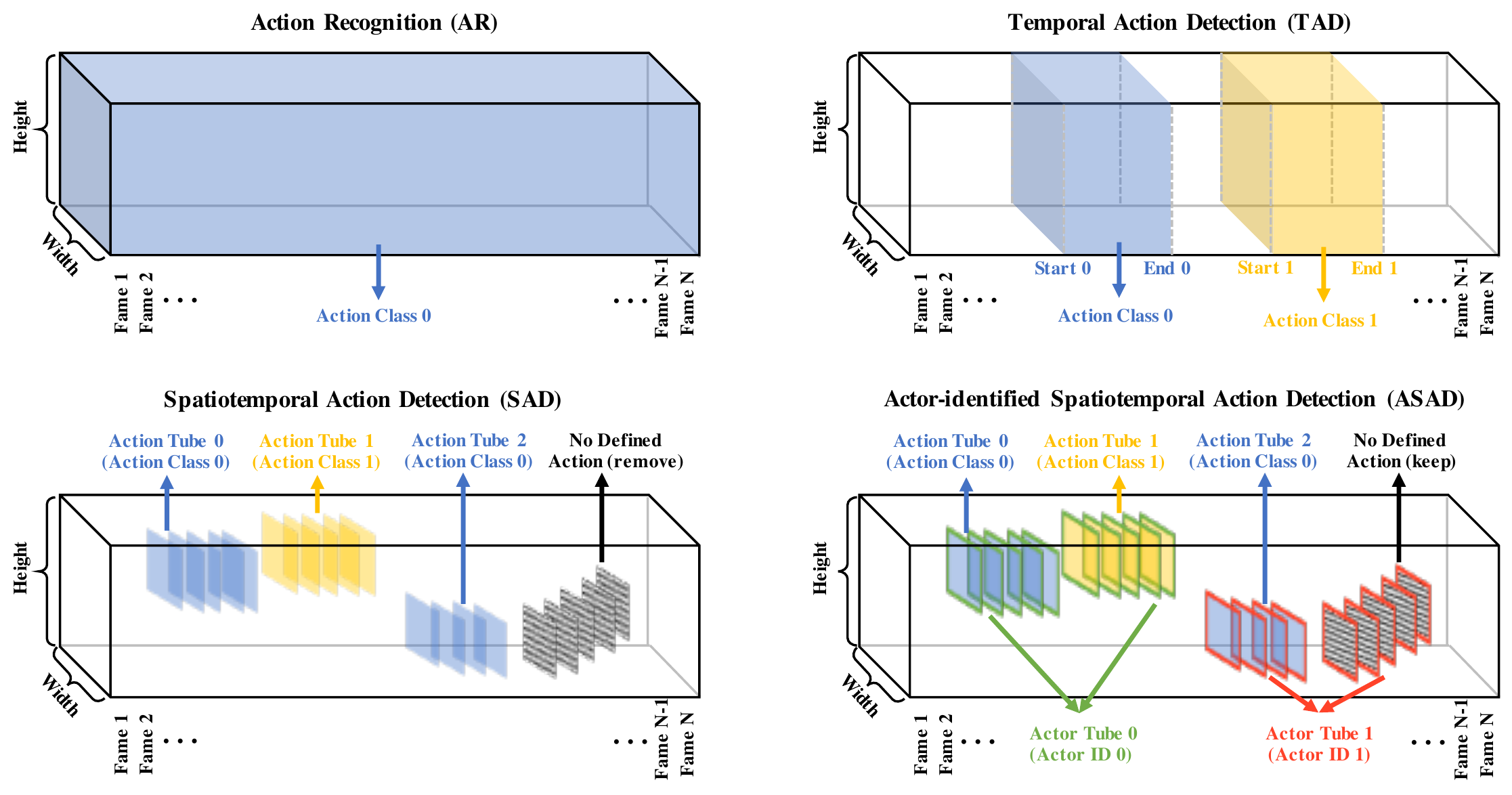} 
  \caption{A comparison of action recognition works, which could be roughly divided into four categories: Action Recognition (AR), Temporal Action Detection (TAD), Spatiotemporal Action Detection (SAD), and our defined Actor-identified Spatiotemporal Action Detection (ASAD). \textbf{Existing works (\ie, AR, TAD, and SAD) ignore to identify actors while our ASAD addresses this issue}.
  Parts of this graph credit to \cite{hutchinson2020video}.}
  \label{fig:ASAD_to_others}
\end{figure*}

Since ASAD is a new task, it poses many new challenges that cannot be addressed by existing methods: i) no dataset is specifically created for ASAD, ii) no evaluation metrics are designed for ASAD, iii) current MOT performance could be the bottleneck to obtain satisfactory ASAD results. To address those problems, we contribute to i) annotate a new ASAD dataset, ii) propose ASAD evaluation metrics by considering multi-label actions and actor identification, iii) improve the data association strategies in MOT to boost the MOT performance, which leads to better ASAD results \footnote{We have organized our proposed dataset and evaluation metrics scripts at GitHub, it will be released.}.

We summarize the main contributions as follows.
\begin{itemize}
  \item We raise a new study task of video action recognition --- Actor-identified Spatiotemporal Action Detection (ASAD). As far as we are aware, it has great importance but has been historically overlooked. ASAD bridges the gap between the existing Spatiotemporal Action Detection (SAD) study and the new demand for identifying actors.
  
  \item We specifically provided a novel dataset for the ASAD study. It covers a rich action category and actor identities.
  
  \item We presented novel metrics for ASAD evaluation. To the best of our knowledge, existing metrics cannot be applied to ASAD, and we are the first to introduce such metrics.
\end{itemize}

\section{Related Works}
\label{sec:related_works}

\section{Video Action Recognition}
In general, video action recognition research can be divided into several categories (Figure~\ref{fig:ASAD_to_others}). Normal Action Recognition (AR) takes an entire video, or, a video clip, as the input and generates a corresponding action class. It is used to understand the overall video concept without specifying the details in the spatial domain and temporal domain. Temporal Action Detection (TAD) gives temporal details to AR, by clarifying the start and end times of an action. Accordingly, one video could be segmented into several temporal components in TAD. Compared with TAD, Spatiotemporal Action Detection (SAD) not only detects the action boundary in the temporal domain but also locates the actor with bounding boxes (or instance masks) in the spatial domain. We generally call such a spatiotemporal boundary the action tube. In this work, we propose Actor-identified Spatiotemporal Action Detection (ASAD) from SAD, by incorporating the unique identity of each actor. 

We summarize the related datasets and studies for AR, TAD, SAD, and ASAD in Table~\ref{tab:ar_works}. To link bounding boxes to action tubes, Multiple Object Tracking (MOT) \cite{yang2021remot} is also commonly applied in SAD. Some SAD works can also track the actor and assign them with unique IDs. However, \emph{based on the evaluation protocol of SAD, the annotation of actor identity may not be provided and the actor identification has not been evaluated}. That means, there is no clear boundary between ASAD and SAD in terms of the method, their difference lies more in the data annotation and evaluation protocols. In detail, the action tube ID given in SAD may not be consistent with actor ID. For example, after the same actor changes his/her action, the corresponding action tube ID changed but the actor ID should remain the same. Unfortunately, such kind of actor ID is not available in most SAD datasets.

As we suppose that MOT and AC are two important components in ASAD, we take a look into the role of MOT and AC in AR, TAD, SAD, and ASAD (see Table~\ref{tab:MOT_AC_roles}). The AC could be a necessary module for all action recognition categories. In SAD, MOT might be used (\eg, on UCF101 + ROAD dataset~\cite{singh2017online}), but not be necessary (\eg, on AVA dataset~\cite{gu2018ava}). However, both MOT and AC are needed in ASAD.

In addition, previous studies~\cite{satoh1999towards,xu2010cast2face,Huang_2018_CVPR} focus on only identifying actors in videos, but without detecting their actions. In this manner, as a new task, ASAD has bridged the gap between the SAD and the actor identification (Table~\ref{tab:SAD_ActorID_works}).

\tabcolsep=3pt
\begin{table*}[h!]
\resizebox{0.9\linewidth}{!}{
\begin{tabular}{c|c|c}
\toprule
Action Recognition Category & Available Datasets & Related Works  \\ \toprule
AR & \makecell[c]{HMDB~\cite{kuehne2011hmdb}, UCF101~\cite{soomro2012ucf101},Sports-1M~\cite{karpathy2014large}, Kinetics-700~\cite{carreira2019short}} & \makecell[c]{\cite{masoud2003method,liu2010human,han2010discriminative,simonyan2014two,tran2015learning,feichtenhofer2016convolutional,liu20163d,yuan2016action,feichtenhofer2019slowfast,gu2020multiple}}
\\ \midrule
TAD & \makecell[c]{ActivityNet~\cite{caba2015activitynet}, YouTube-8M~\cite{abu2016YouTube},THUMOS~\cite{idrees2017thumos}, HACS~\cite{zhao2019hacs} } & \makecell[c]{\cite{soomro2015action,wang2016temporal,xu2017r,GaoYN17,chao2018rethinking,lin2019bmn}}
\\ \midrule
SAD & \makecell[c]{UCF101+ROAD~\cite{singh2017online}, DALY~\cite{weinzaepfel2016human}, Hollywood2Tubes~\cite{mettes2016spot} AVA~\cite{gu2018ava}, \\AVA-Kinetics~\cite{li2020ava}, ActEV~\cite{yooyoung2019actev18}} &\makecell[c]{\cite{gkioxari2015finding,weinzaepfel2015learning,zhu2017tornado,hou2017tube,singh2017online,kalogeiton2017action,yang2019framework}\\
\cite{girdhar2018better,yang2019step,ulutan2020actor,pan2020actor,li2020actions,li2020cfad,tang2020asynchronous}}
\\ \midrule
ASAD & \makecell[c]{Okutama~ \cite{barekatain2017okutama} (available but not ideal)} &
\textbf{Ours}
\\\bottomrule
\end{tabular}
}
\caption{The related datasets and studies for AR, TAD, SAD, and ASAD. Note that, unlike other SAD datasets, actor ID is given in annotations of Okutama~\cite{barekatain2017okutama}
but the ASAD evaluation has not been explored. Besides, the Okutama dataset consists of 4K-resolution drone videos, which may only cover very limited scenarios of ASAD. In addition, \textbf{some SAD models, such as ROAD~\cite{singh2017online}, AlphAction~\cite{tang2020asynchronous}, and ACAM~\cite{ulutan2020actor}, may potentially generate ASAD results but were evaluated by the SAD protocol in the original works. That is, the consistency of actor identity is ignored in these works.}}
\label{tab:ar_works}
\end{table*}

\tabcolsep=3pt
\begin{table*}[h!]
\centering
\resizebox{0.75\linewidth}{!}{
\begin{tabular}{c|c|c}
\toprule
            Approaches & \makecell[c]{Identifying Actors} & \makecell[c]{Detecting Actions}  \\ \midrule
            \makecell[c]{Spatiotemporal Action Detection (SAD)\\
            \cite{gkioxari2015finding,weinzaepfel2015learning,zhu2017tornado,hou2017tube,singh2017online,kalogeiton2017action,yang2019framework,girdhar2018better,pan2020actor,li2020actions,li2020cfad,tang2020asynchronous}} & \ding{55}& \ding{51} \\ \midrule
            Actor Identification~\cite{satoh1999towards,xu2010cast2face,Huang_2018_CVPR} & \ding{51}& \ding{55} \\ \midrule
            Actor-identified Spatiotemporal Action Detection (ASAD) & \ding{51}& \ding{51}
            \\\bottomrule
\end{tabular}
}
\caption{A comparison of SAD, Actor Identification, and ASAD.}
\label{tab:SAD_ActorID_works}
\end{table*}

\tabcolsep=3pt
\begin{table*}[h!]
\centering
\resizebox{0.6\linewidth}{!}{
\begin{tabular}{c|c|c}
\toprule
Action Recognition Category & Using MOT & Using AC  \\ \toprule
Action Recognition (AR) & Not Need & Need
\\ \midrule
Temporal Action Detection (TAD) & Not Need & Need
\\ \midrule
Spatiotemporal Action Detection (SAD) & May Not Need & Need
\\ \midrule
Actor-identified Spatiotemporal Action Detection (ASAD) & Need & Need
\\\bottomrule
\end{tabular}
}
\caption{The role of MOT and AC in AR, TAD, SAD, and ASAD. For some evaluation protocols of SAD, there is no need to link detection to tubes and MOT may not be used.}
\label{tab:MOT_AC_roles}
\end{table*}

\tabcolsep=3pt
\begin{table*}[h!]
\centering
\resizebox{0.7\linewidth}{!}{
\begin{tabular}{c|c|c}
\toprule
            Approaches & \makecell[c]{Online Data \\ Association} & \makecell[c]{Offline Data \\Association}  \\ \midrule
            \makecell[c]{\cite{qu2007real,Bewley2016_sort,Wojke2017SimpleOA,Bochinski2017,dai2018instance,li2019joint,fu2019multi,jiang2019graph,ma2019deep,wang2019towards,li2020graph,braso2020learning,Weng_2020_CVPR,zhou2020tracking,zhang2020simple,fernando2018tracking,chang2019argoverse}}  & \ding{51}& \ding{55} \\ \midrule
            \makecell[c]{\cite{zhang2008global,wen2014multiple,zhang2015multi,tang2017multiple,ristani2018features,Ma2018CustomizedMT,lifted_disjoint_paths_2020_ICML} } & \ding{55}& \ding{51} 
            \\\bottomrule
\end{tabular}
}
\caption{Online and offline MOT methods.}
\label{tab:online_offline_MOT}
\end{table*}

\section{Multiple Object Tracking}

Since Multiple Object Tracking (MOT) plays an important role in Actor-identified Spatiotemporal Action Detection (ASAD), we further provide an overview of MOT-related works.

Generally, MOT is applied to connect identical observations into tracklets based on the similarity of MOT features. Specifically, MOT methods can be divided into two categories --- online MOT and offline MOT (Table~\ref{tab:online_offline_MOT}.). The online data association is performed on observations that are available up to the current frame. Different from online approaches, offline data association takes global observations into consideration, which may not be applicable for real-time applications but be ideal for assisting annotation works. Numerous offline approaches have been proposed in previous studies~\cite{luo2014multiple}. Among them, formulating MOT data association as a global clustering problem has achieved great successes~\cite{yang2020remots}

\section{Action Classification}

\begin{figure*}[th!]
\centering
  \includegraphics[width=0.65\linewidth]{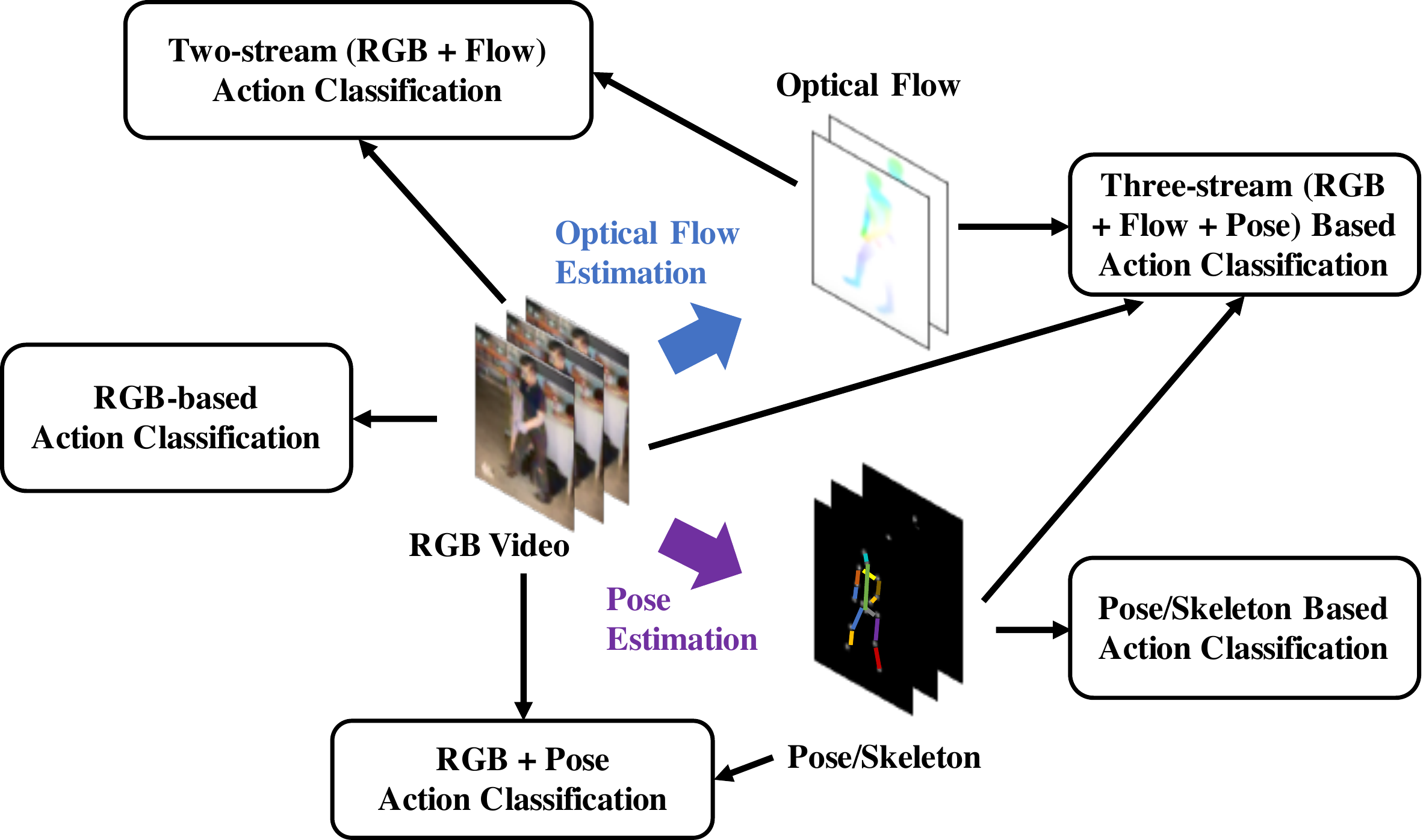}
   \caption{Categories of Action Classification (AC) models.}
   \label{fig:AC_cat}
\end{figure*}

The Action Classification (AC) model plays such a role to map the spatiotemporal information to action categories. There are numerous AC studies considering the approaches of utilizing features and designing the model structure. In detail, Action Classification (AC) approaches could be divided into 5 categories, including RGB AC, RGB + Flow AC, Pose AC, RGB + Pose AC, and RGB + Flow + Pose AC, as shown in Figure~\ref{fig:AC_cat}. Based on these 5 categories, we list the corresponding studies in Table~\ref{tab:ac_works}. 

\tabcolsep=3pt
\begin{table*}[h!]
\resizebox{0.8\linewidth}{!}{
\begin{tabular}{c|c|c|c|c|c}
\toprule
Approaches & RGB AC & RGB + Flow AC  & Pose AC & RGB + Pose AC & RGB + Flow + Pose AC\\ \midrule
Large-scale AC~\cite{karpathy2014large} & \ding{51} & \ding{55} & \ding{55} & \ding{55}  & \ding{55} \\ \midrule
Two-Stream \cite{simonyan2014two}  & \ding{55} & \ding{51} & \ding{55} & \ding{55} & \ding{55} \\ \midrule
\makecell[c]{C3D~\cite{tran2015learning}, I3D~\cite{carreira2017quo}, ECO~\cite{zolfaghari2018eco},\\ P3D~\cite{qiu2017learning},
FastSlow~\cite{feichtenhofer2019slowfast}}
& \ding{51} & \ding{51} & \ding{55} & \ding{55}  & \ding{55}\\ \midrule
\makecell[c]{HCN~\cite{li2018co},
2s-AGCN~\cite{shi2019two}, \\DD-Net~\cite{yang2019make}} & \ding{55} & \ding{55} & \ding{51} & \ding{55}  & \ding{55}\\ \midrule
Potion~\cite{choutas2018potion}, PA3D~\cite{yan2019pa3d} & \ding{55} & \ding{55} & \ding{55} & \ding{51}  & \ding{55}\\ \midrule
Chained AC~\cite{zolfaghari2017chained} & \ding{55} & \ding{55} & \ding{55} & \ding{55}  & \ding{51}
\\\bottomrule
\end{tabular}
}
\caption{The properties of action classification works}
\label{tab:ac_works}
\end{table*}

\section{Proposed ASAD Dataset and Evaluation Metrics}

Given a video, Actor-identified Spatiotemporal Action Detection (ASAD) aims to detect the spatiotemporal boundaries (\ie, tracklets/actor tubes) for each actor, assign each actor a unique identity, and obtain the actions of actors at each moment. Consequently, the ASAD dataset should include those factors and the ASAD metrics should verify the performance on those factors. 

\subsection{Dataset for ASAD}

\begin{figure}[h!]
	\centering
		\includegraphics[width=\linewidth]{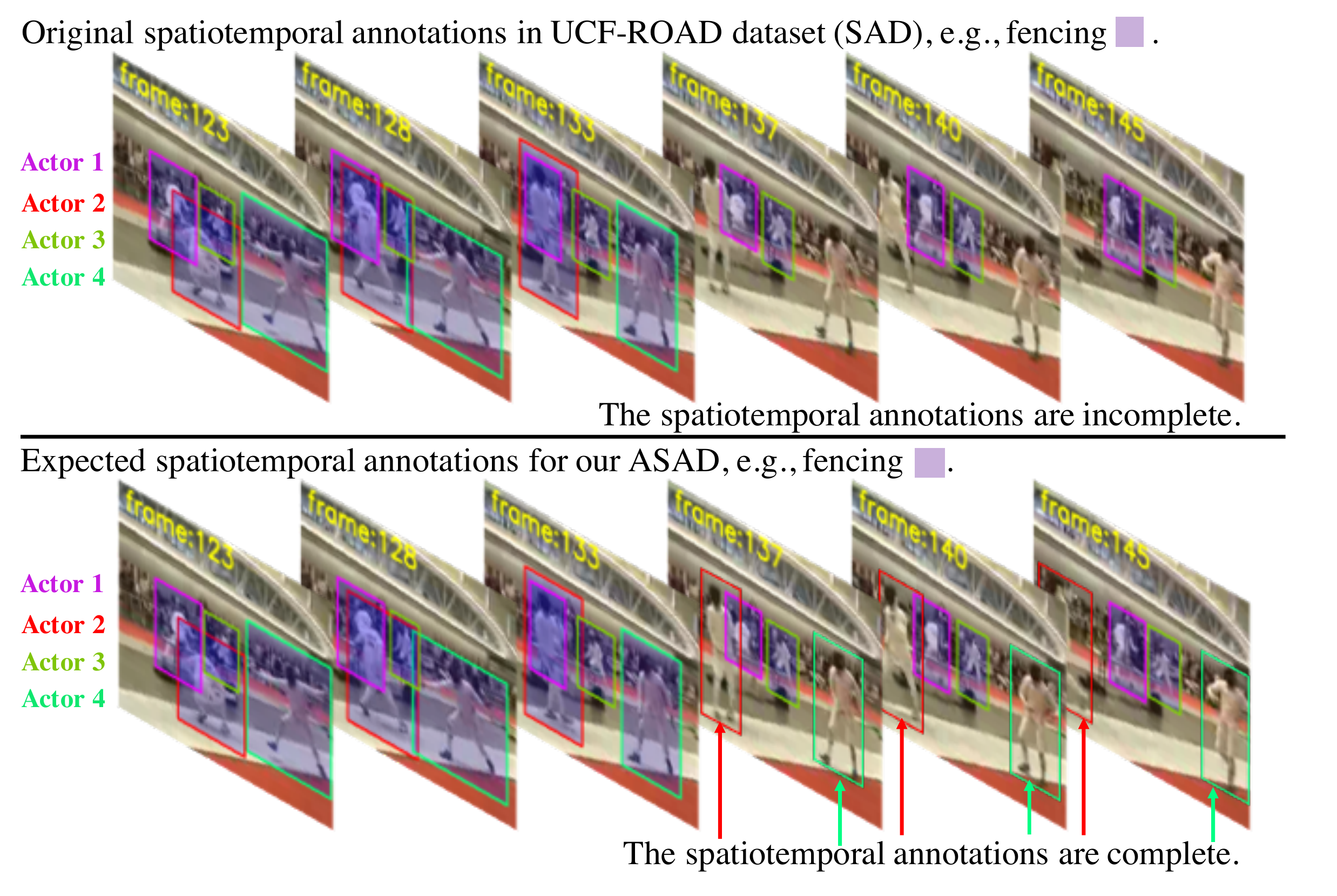}
  \caption{Comparison between SAD and ASAD spatiotemporal annotation by using UCF101+ROAD~\cite{singh2017online} as an example. The annotation in ASAD should complete the spatiotemporal boundary for each actor in the entire video, no matter if the defined action is finished or not.}
  \label{fig:SAD_vs_ASAD_by_UCF}
\end{figure}

\begin{figure}[h!]
	\centering
		\includegraphics[width=\linewidth]{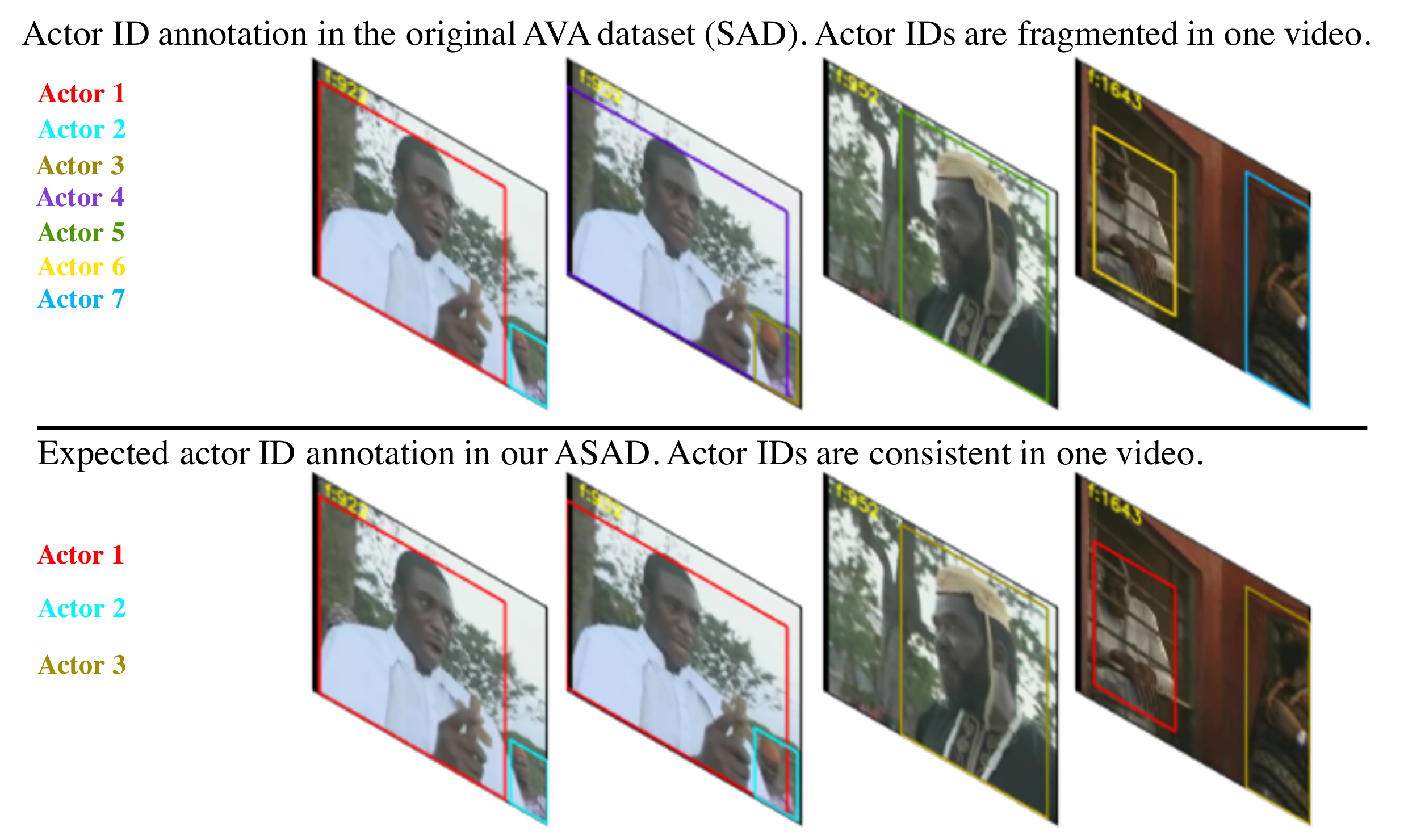}
  \caption{Comparison between SAD and ASAD actor ID annotation by using AVA~\cite{gu2018ava} as an example. In a single video, while the existing SAD dataset may assign multiple actor IDs to the same actor, our ASAD assigns the unique actor ID the actor.}
  \label{fig:SAD_vs_ASAD_by_AVA}
\end{figure}

By reviewing existing action recognition datasets (Section~\ref{sec:related_works}), it can be noticed that a proper ASAD dataset may not be available. Although the existing Spatiotemporal Action Detection (SAD) dataset might be similar to our desired ASAD dataset, the actor identity is not properly annotated in the SAD dataset. We illustrate the annotation difference between SAD and ASAD data annotation by using UCF101+ROAD dataset~\cite{singh2017online} and AVA dataset~\cite{gu2018ava} as examples (Figures~\ref{fig:SAD_vs_ASAD_by_UCF} and \ref{fig:SAD_vs_ASAD_by_AVA}). In the UCF101+ROAD dataset, the spatiotemporal boundaries are incomplete. Since actor identification is not the concern in SAD, after the predefined action is finished, spatiotemporal annotation is not available. In contrast, the annotation in ASAD should complete the spatiotemporal boundary for each actor in the entire video, no matter if the defined action is finished or not. In the AVA dataset, despite the actor IDs being given, multiple actor IDs have been assigned to the same actor in a single video, which is incorrect for actor identification. For actor identification purposes, the unique actor ID should be assigned to each actor in one piece of video. Besides, while some remote surveillance video datasets are equipped with spatiotemporal boundaries, actor identities, and acting classes, they focus on the special scene (\eg, remote surveillance) and may not be suitable for the general ASAD study. For example, Okutama dataset~\cite{barekatain2017okutama}  and PANDA~\cite{wang2020panda} only record tiny scale actors and cover a small group of human daily activities. 

\begin{figure*}[th!]
	\centering
		\includegraphics[width=0.68\textwidth]{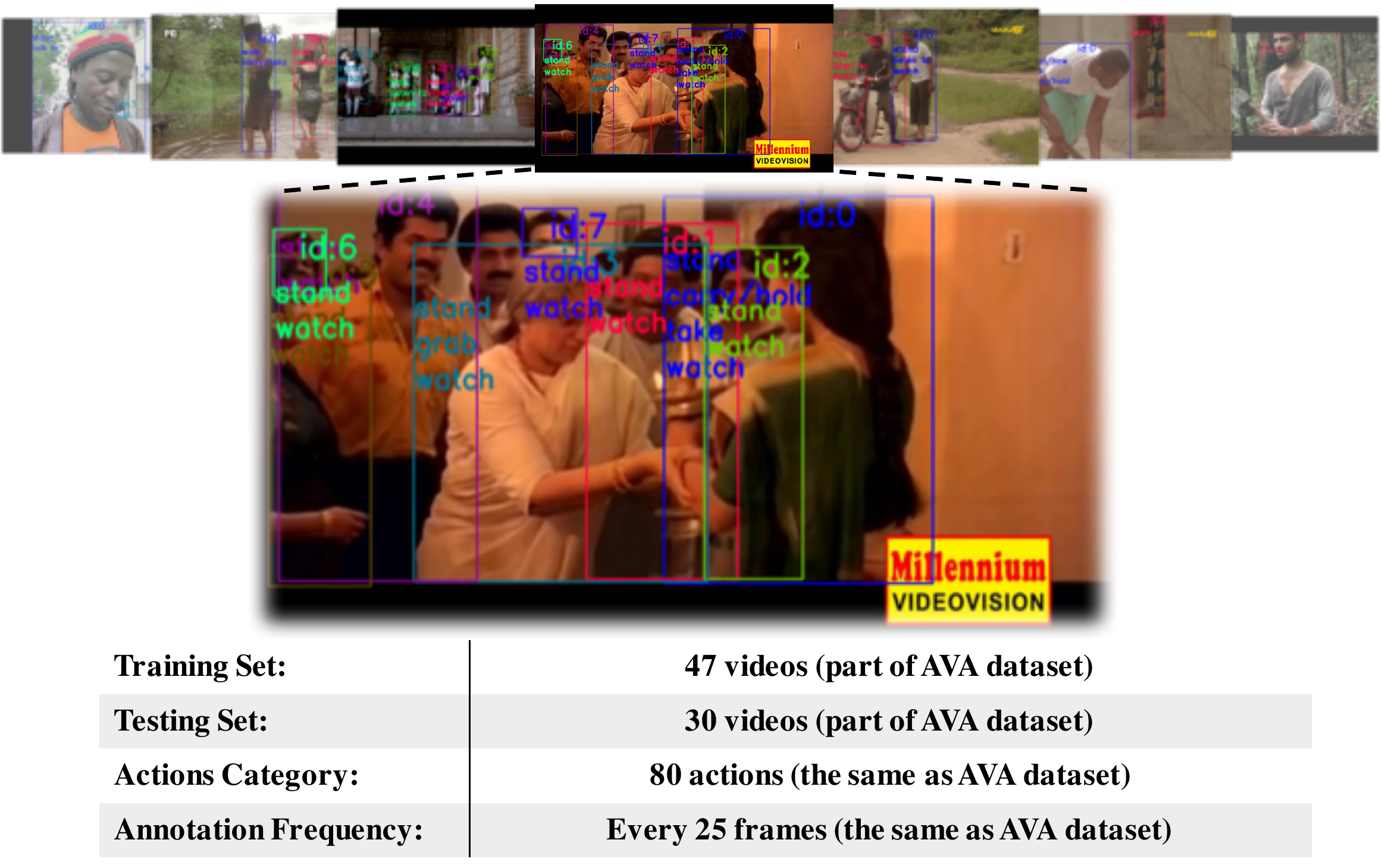}
  \caption{We create a new ASAD dataset based on existing AVA dataset~\cite{gu2017ava}, by assigning the unique actor identity to each actor.}
  \label{fig:a-ava_dataset_summary}
\end{figure*}

\begin{figure*}[h!]
	\centering
		\includegraphics[width=0.8\textwidth]{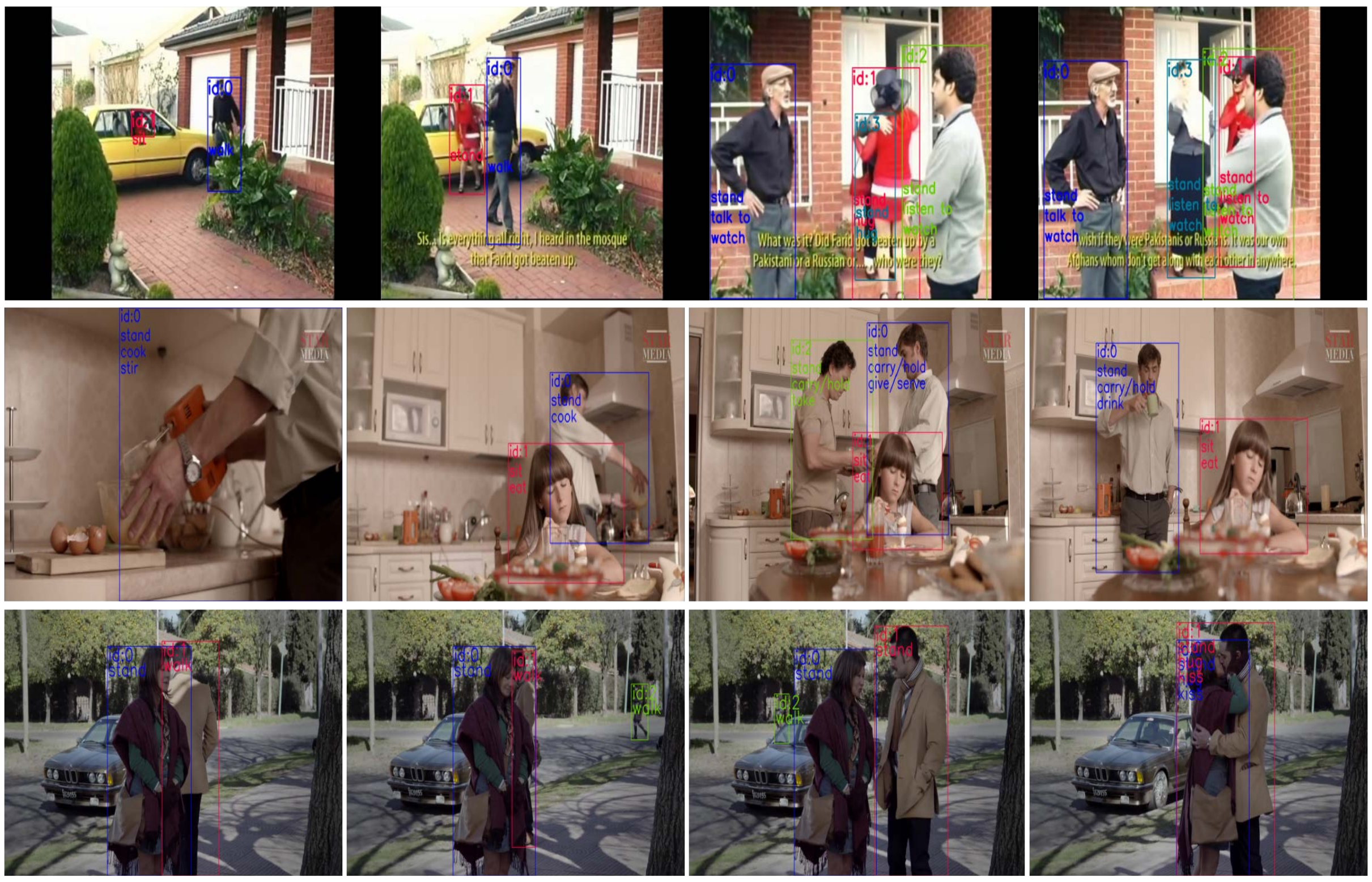}
  \caption{Illustration of our Actor-identified AVA dataset.}
  \label{fig:A-AVA_examples}
\end{figure*}

Due to the above reasons, we are motivated to annotate a new ASAD dataset. Compared with the SAD dataset, the ASAD dataset requires to add correct actor identities. As the AVA dataset~\cite{gu2018ava} is a canonical SAD dataset and TAO dataset~\cite{dave2020tao} offers some actor identity annotations on it, we select a part of the AVA dataset to make an ASAD dataset (Figure~\ref{fig:a-ava_dataset_summary}). Note that, we mainly selected video clips that have visible actors and multiple actors available. Meanwhile, due to the heavy annotation cost, only 77 video clips are selected among 430 AVA video clips. We named our ASAD dataset A-AVA, which represents the Actor-identified AVA dataset. A-AVA dataset contains 47 videos for training and 30 videos for testing. Be the same as the AVA dataset, there are 80 action categories in the A-AVA dataset, and, every 25 frames (\ie, around 1 second), the annotation is given once. In the A-AVA dataset, the spatiotemporal boundaries, actor identities, and corresponding actions are all annotated. More examples are illustrated in Figure~\ref{fig:A-AVA_examples}. 

\begin{figure*}[h!]
	\centering
		\includegraphics[width=0.9\textwidth]{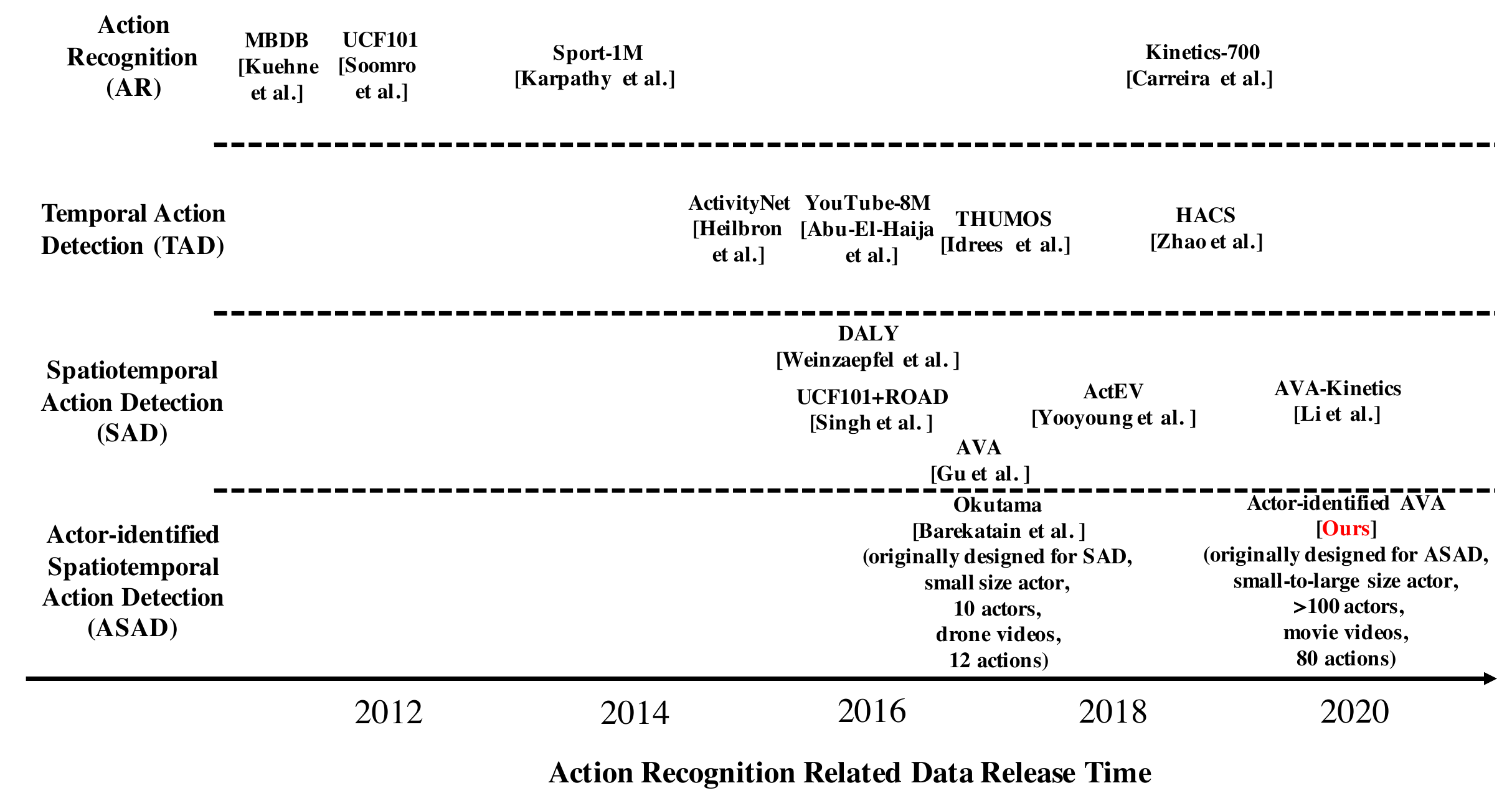}
  \caption{A historical timeline overview of datasets intended for video action recognition studies.}
  \label{fig:dataset_timeline}
\end{figure*}

We present the historical role of our A-AVA dataset in Figure~\ref{fig:dataset_timeline}. As the first dataset that is specifically designed for the ASAD study, the A-AVA dataset covers a rich diversity of video scenes, as indoor and outdoor, different times of the day, various actor scales, and more. Those properties are not available in the previous dataset (\ie, Okutama dataset). A-AVA dataset has bridged the gap between the SAD dataset and actor identification dataset.

\subsection{Evaluation Metrics for ASAD}
\begin{figure*}[h!]
	\centering
		\includegraphics[width=0.68\textwidth]{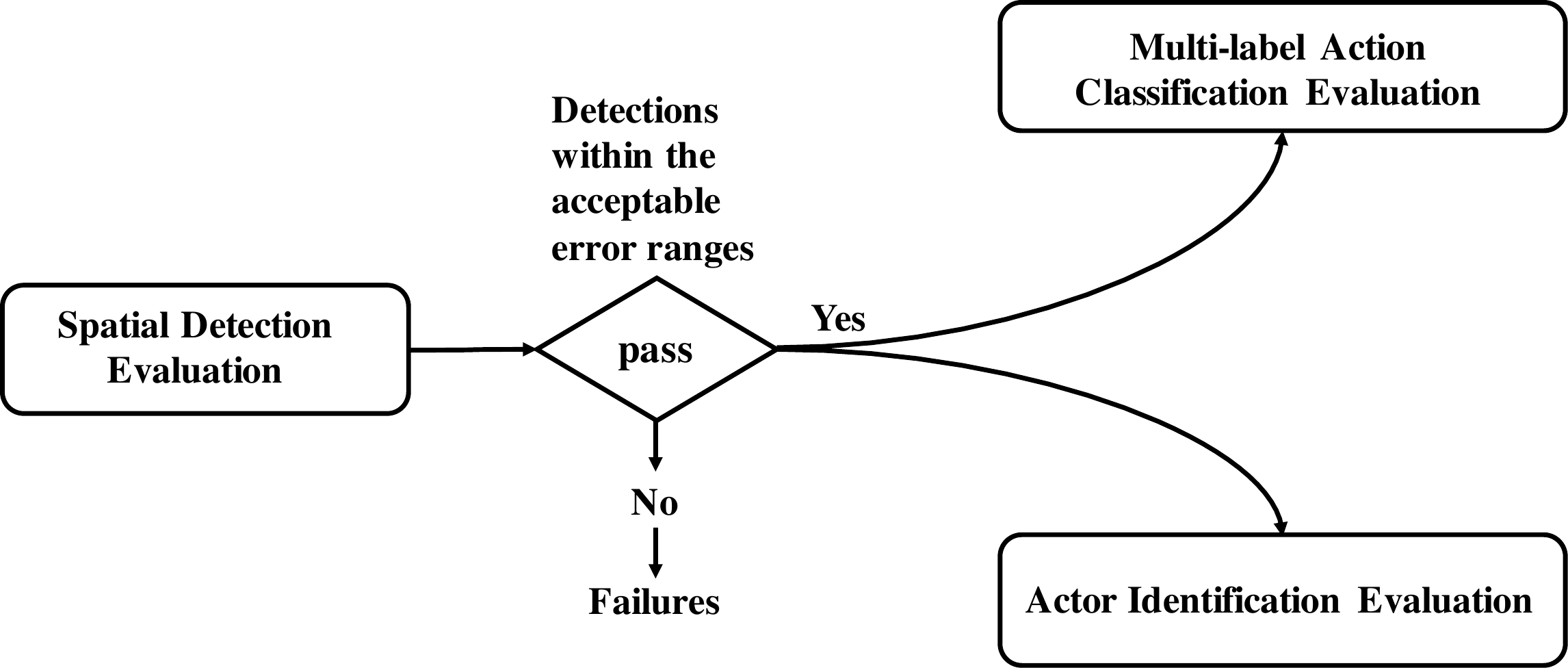}
  \caption{Overview of our ASAD metrics, which evaluate the performance of spatial detection, action classification, and actor identification.}
  \label{fig:ASAD_metrics}
\end{figure*}

When considering the multi-label action, ASAD evaluation could be a complicated task. Unlike single-label SAD evaluation~\cite{gu2017ava}, it is challenging to simultaneously evaluate multi-label action classification and actor identification with spatial detection. To address this issue, we suggest evaluating ASAD from three aspects and then consider their overall performance. The three aspects include Spatial Detection Evaluation, Actor Identification Evaluation, and Multi-label Action Classification Evaluation (Figure~\ref{fig:ASAD_metrics}). 

\subsection{Spatial Detection Evaluation}
We take the object detection metrics~\cite{everingham2010pascal,lin2014microsoft,hosang2015makes} to evaluate the spatial detection performance. First, we calculate Intersection over Union (IoU), which is defined by 
\begin{equation}
\label{eq:IoU}
IoU = \frac{bbox^{pred}\cap bbox^{true} }{bbox^{pred}\cup bbox^{true} }
\end{equation}
where $bbox^{pred}$ and $bbox^{true}$ represent the predicted bounding box and the ground-truth box, respectively.

Second, based on the IoU value, True Positive (TP), False Positive (FP), and False Negative (FN) are defined by
\begin{itemize}
\item True Positive (TP): A correct detection with an IoU greater the threshold.
\item False Positive (FP): A wrong detection with an IoU smaller than the threshold.
\item False Negative (FN): A ground truth not detected.
\end{itemize}
and the corresponding Precision and Recall are
\begin{equation}
\label{eq:precision_recall}
\begin{split}
Precision &= \frac{TP}{TP + FP} \\
Recall &=  \frac{TP}{TP + FN}
\end{split}
\end{equation}

By traversing through all thresholds for detection confidence, different pairs of precision and recall can generate the precision-recall curve, which indicates the association between precision and recall. To reduce the effect of the wiggles in the curve, the precision-recall curve is interpolated as $p_{interp}$. The $p_{interp}$ at recall score $r$ is assigned with the highest precision for $r>r^{'}$:
\begin{equation}
\label{eq:P}
p_{interp}(r) = \max_{r>r^{'}} p(r^{'})
\end{equation}

Since we only treat humans as the actor, there is only one class for the detection, and therefore we utilize Average Precision (AP), other than Mean Average Precision (mAP), for the spatial detection evaluation. AP is the area under the interpolated precision-recall curve, which can be calculated using the following formula:
\begin{equation}
\label{eq:AP}
AP = \sum^{n-1}_{i=1} (r_{i+1} - r_{i}) p_{interp}(r_{i+1})
\end{equation}

In this thesis, we assume any spatial detection with an IoU value larger than 0.5 is True Positive, and the corresponding metrics are represented as AP@0.5.

\subsection{Actor Identification Evaluation}

While actor classification has the pre-defined actor identities, actor identification assigns each actor a unique identity and the number of actor identities is non-parametric. Therefore, we utilized part of Multiple Object Tracking (MOT) evaluation metrics for actor identification evaluation, as IDF1 (ratio of correctly identified detections), MT (mostly tracked targets), ML (mostly lost targets), and ID Switches. Those identification metrics were introduced by \cite{ristani2016performance, milan2016mot16} and have been popularly utilized for a while. More specifically, the IDF1, MT, and ML are respectively defined by
\begin{equation}
\label{eq:IDF1}
IDF1 = \frac{2 IDTP}{2 IDTP + IDFP + IDFN}
\end{equation}
where IDTP, IDFP, IDFN respectively represent the True Positive ID, the False Positive ID, the False Negative ID.

\begin{equation}
\label{eq:MT_ML}
\begin{split}
MT = \sum_{i \in N_{\mathcal{T}_{true}}} \mathbb{1} \{ \frac{len(\mathcal{T}^{pred}_{i})}{len(\mathcal{T}^{true}_{i})} \geqslant 0.8 \} \\
ML = \sum_{i \in N_{\mathcal{T}_{true}}} \mathbb{1} \{ \frac{len(\mathcal{T}^{pred}_{i})}{len(\mathcal{T}^{true}_{i})} \leqslant  0.2 \}
\end{split}
\end{equation}
where $\mathcal{T}^{pred}_{i}$ and $\mathcal{T}^{true}_{i}$ respectively denote the predicted and the ground-truth Tracklet $i$, the number of $\mathcal{T}^{true}_{i}$ is $N_{\mathcal{T}_{true}}$. If the prediction matches for the ground truth more than $80\%$ of its life span, it is regarded as mostly tacked (MT). If the prediction only matches for the ground truth less than $20\%$ of total length, it is regarded as mostly lost (ML).

\subsection{Multi-label Action Classification Evaluation}

It is intuitive to consider that each actor could take several actions simultaneously, which are corresponding to multi-label actions. For instance, an actor could be making a phone call and walking at the same time. Due to the lack of evaluation metrics, conventional Action Recognition studies have been evaluated with only the single-label action for a while~\cite{barekatain2017okutama,gu2017ava}. Therefore, we provide metrics for multi-label ASAD evaluation, which considers the evaluations of multi-label multi-class action classification and actor identification.

The evaluation metrics for actor detection and multi-label classification have been well-studied separately~\cite{everingham2010pascal,gibaja2015tutorial}, but the problem remains on how to associate them together for multi-label ASAD evaluation. 

A simple approach could be evaluating the ``actor'' actor detection performance for all detected samples and then evaluating the multi-label action recognition performance for positively detected samples. For instance, assuming that a predicted sample is positive when IoU $\geq 0.5$ for the predicted and ground-truth bounding boxes, we can apply $HL\text{@}0.5$, which corresponds to Hamming Loss associated with IoU $\geq 0.5$, to measure its multi-label classification performance. 

\begin{figure*}[htb!]
\centering
  \includegraphics[width=0.9\linewidth]{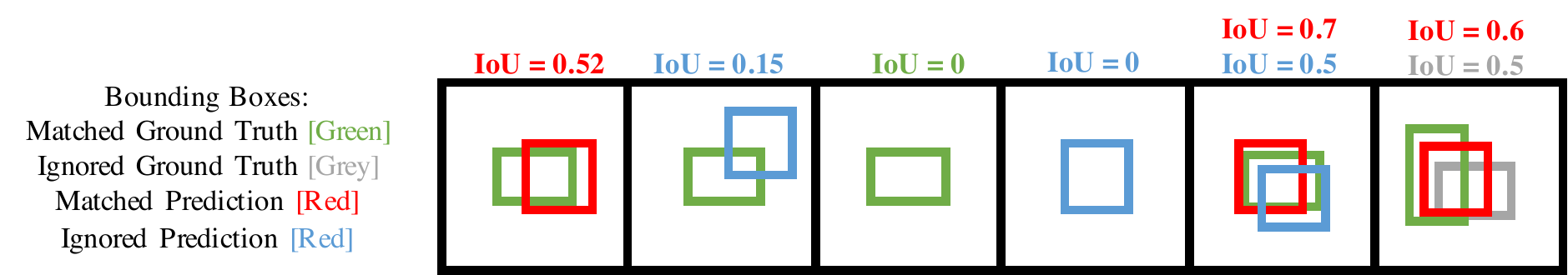}
  \caption{Illustration of matching pair between the ground-truth and the predicted samples.}
  \label{fig:match_pairs}
\end{figure*}

Note that, due to the object occlusions, the IoU value between multiple actors could be larger than $0.5$. To remove such ambiguity, we apply the Hungarian Algorithm for bipartite matching between the predicted bounding boxes and the ground-truth bounding boxes before comparing their classification results. Meanwhile, a pair that has IoU $<0.5$ will be excluded before calculating their Hamming Loss. We illustrate these cases in Figure~\ref{fig:match_pairs}.

In detail, we utilize matrix $\mathcal{D}_{i,j}$ to represent the matching distance between each ground-truth bounding box (denoted by $i$) and predicted bounding box (denoted by $i$), and we obtain $\mathcal{D}_{i,j}$ by 
\begin{equation}
\label{eq:iou_dist}
\begin{split}
\mathcal{D}_{i,j} =
\left\{\begin{matrix}
&1,  &\text{if}\ IoU_{i,j}<0.5; \\
&1-IoU_{i,j},  &\text{otherwise}.
\end{matrix}\right .
\end{split}
\end{equation}

Next, we employ linear assignment~\cite{crouse2016implementing} to obtain the optimal assignment $\mathcal{M}^*$ with
\begin{equation}
\label{eq:hungarian_2}
\mathcal{M}^* = \argmin_{\mathcal{M}} \sum _{i}\sum _{j}\mathcal{D}_{i,j}\mathcal{M}_{i,j},
\end{equation}
where $\mathcal{M}$ is a Boolean matrix. When the row $i$ (\ie, ground truth box $i$) is assigned to column $j$ (\ie, predicted box $j$), we have $\mathcal{M}_{i,j}=1$. Each row can be assigned to at most one column and each column to at most one row.

Since matching pairs that have IoU value less than 0.5, we further process $\mathcal{M}^*$ by
\begin{equation}
\label{eq:remove_iou<0.5}
\begin{split}
\mathcal{M}^*_{i,j} = 
\left\{\begin{matrix}
&0,  & \text{if}\ \mathcal{D}_{i,j} = 1; \\
&\mathcal{M}^*_{i,j},  &\text{otherwise}.
\end{matrix}\right .
\end{split}
\end{equation}

Referring to $\mathcal{M}^*$, we select matched pairs to evaluate the corresponding action labels with Hamming Loss. The number of matching pairs are represented by $N_{actors\text{@}0.5}$ (\ie, $\mathcal{M}^*=1$). Below, we show how the $HL\text{@}0.5$ is extended from the original Hamming Loss. 
\begin{equation}
HL\text{@}0.5 = 
\frac{1}{N_{actors\text{@}0.5}} \frac{1}{N_{labels}}\sum_{i=1}^{N_{actor\text{@}0.5}}
\sum_{l=1}^{N_{labels}}Y_{true}^{i,l}~\textbf{XOR}~ Y_{pred}^{i,l}~,
\end{equation}
where $\textbf{XOR}$ is an exclusive-or operation and $N_{labels}$ stands for the number of action categories. $Y_{true}$ and $Y_{pred}$ are boolean arrays that denote the ground truth and predicted labels, respectively. 

\section{Experiment}
We set up the first benchmark for the ASAD study. Experiments are conducted on our A-AVA dataset and evaluated by our ASAD metrics.

\subsection{ASAD Framework}
As we have discussed in related works, some SAD models, such as ROAD~\cite{singh2017online}, AlphAction~\cite{tang2020asynchronous}, and ACAM~\cite{ulutan2020actor}, are consist of Multiple Object Tracking (MOT) and Action Classification (AC) modules. They could generate ASAD results but were evaluated by the SAD protocol in the original works. Based on the evaluation protocol of SAD, the annotation of actor identity may not be provided and the actor identification has not been evaluated. In other words, there is no clear boundary between ASAD and SAD in terms of the method, their difference more lies in the data annotation and evaluation protocols.

\begin{figure*}[h!]
	\centering
		\includegraphics[width=0.7\textwidth]{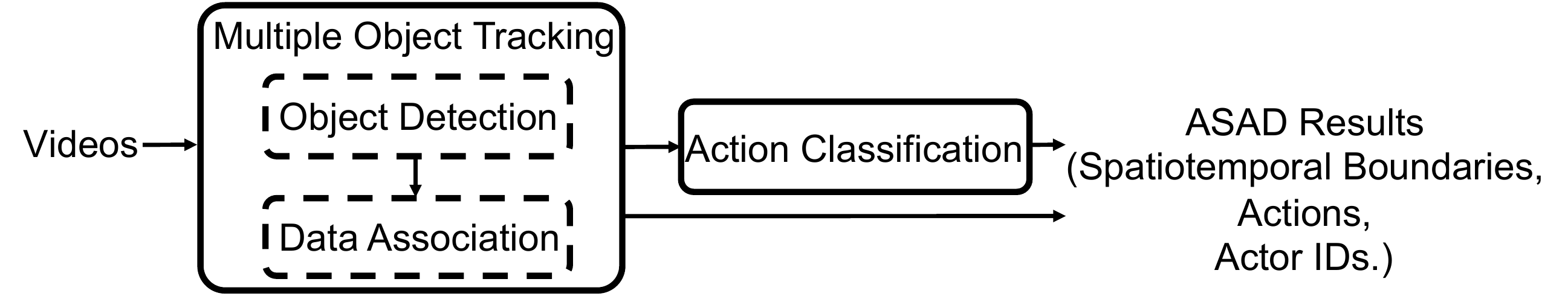}
  \caption{Overview of the basic ASAD framework.}
  \label{fig:ASAD_framework}
\end{figure*}

Without changing the basic structure, letting the above SAD methods to output actor identities with their original outputs can make ASAD frameworks. In this study, we let the off-the-shelf SAD methods to output actor identities that are generated by their MOT module. In this manner, they can perform as ASAD frameworks. In Figure~\ref{fig:ASAD_framework}, we summarize the basic structure of ASAD framework that is adapted from SAD models. Generally, an ASAD framework takes RGB videos as the input and outputs the bounding boxes, unique actor identity, and actions of each actor.

\subsection{Experiment Results}

By using our A-AVA dataset, the performance of spatiotemporal detection, action classification, and actor identification can be jointly evaluated. To generate a better actor identification result, we may need to focus on the data association strategy in MOT. We first evaluated our ASAD baseline with online MOT methods~\cite{Wojke2017SimpleOA}. We showed the result of using our ASAD evaluation metrics in Table~\ref{tab:ASAD_result_2}. It can be noticed that the action identification performance is unsatisfactory. Using online MOT methods becomes the bottleneck to obtain satisfactory ASAD results. We then replaced the online MOT module to be an offline MOT method introduced in \cite{yang2021remot}. Consequently, using an offline MOT module led to better actor identification result: the offline MOT module gave a further gain in IDF1 and ML over the online MOT module, and also reduced the ML and ID Switches.

\tabcolsep=3pt
\begin{table*}[h!]
\centering
\resizebox{0.8\linewidth}{!}{
\begin{tabular}{l|c|c|c|c|c|c}
\toprule
\multirow{2}{*}{Approaches} & \multicolumn{1}{c|}{Actor Detection Evaluation} & \multicolumn{1}{c|}{Action Classification Evaluation} &
\multicolumn{4}{c}{Actor Identification Evaluation} \\ \cmidrule(r){2-7}
 & AP@0.5 (\%)$\uparrow$ & HL@0.5 (0$\sim$1)$\downarrow$ & IDF1 (\%) $\uparrow$ & MT (\%)$\uparrow$ & ML (\%)$\downarrow$ & \# ID Sw.$\downarrow$ \\ \midrule
\makecell[l]{ASAD Baseline \\ w/ online MOT} & 72.4 &0.06 & 60.4 & 67.3& 10.5 & 413\\ \hline
\makecell[l]{ASAD Baseline  \\ w/ offline MOT} & 72.4 & 0.06 & 71.4 & 88.4 &5.2 & 273
\\\bottomrule
\end{tabular}
}
\caption{Comparison of using different MOT modules in ADAD frameworks. We utilize our A-AVA dataset and ASAD evaluation metrics, where $\uparrow$($\downarrow$) indicates that the larger(smaller) the value is, the better the performance.}
\label{tab:ASAD_result_2}
\end{table*}

\subsection{Discussion}
\begin{figure*}[h!]
	\centering
		\includegraphics[width=0.8\textwidth]{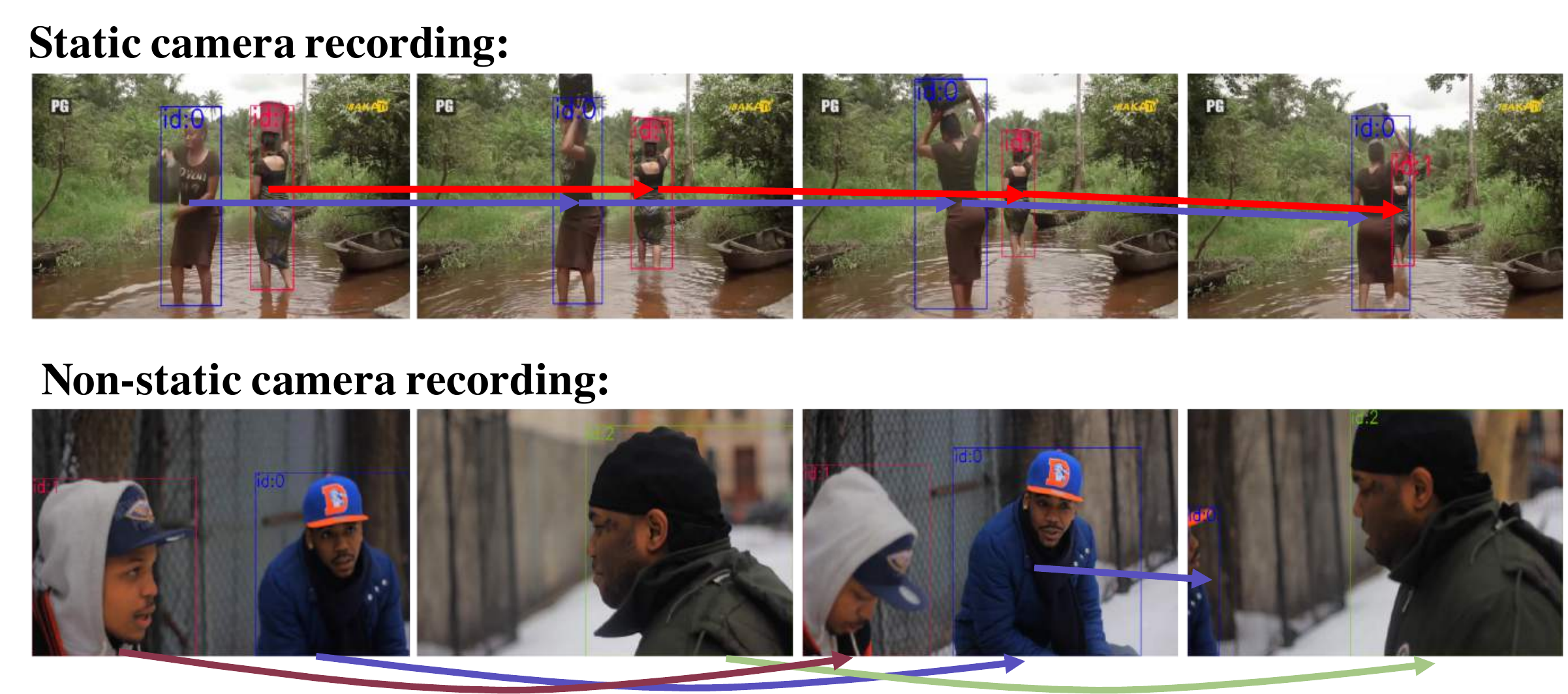}
  \caption{The difference of motion consistency in static camera recording videos and non-static camera recording videos.}
  \label{fig:static_moving_camera}
\end{figure*}

Why does offline MOT have better performance in our A-AVA dataset in terms of actor identification? For static camera recording, motion consistency is an important cue for data association. In contrast, for non-static camera recording, the motion consistency assumption could be failed. Whenever the viewpoint suddenly changes in videos, it is challenging to track the correct actor identities. This issue frequently happens in the movies and phone-recorded videos (Figure~\ref{fig:static_moving_camera}). Most online MOT methods (\eg, \cite{Wojke2017SimpleOA}) may have an over-reliance on the motion consistency and therefore cause failure cases in our A-AVA dataset. Employing offline MOT alleviates this issue by determining the correspondence between observations more by their appearance similarity. Moreover, applying an offline MOT solution utilizes the global information to further reduce ID switches and generate robust actor identification results.

\section{Conclusion}

In this paper, we introduced a novel task, Actor-identified Spatiotemporal Action Detection (ASAD), which marks the first effort in the computer vision community to jointly study spatiotemporal boundaries, actor identities, and corresponding actions. ASAD is ideal for action recognition applications when multiple actors are included, such as Human-computer Interaction, basketball/soccer games, and grocery operations monitoring, etc. To study ASAD, we are excited to offer a corresponding A-AVA dataset. As the first dataset that is specifically designed for the ASAD study, the A-AVA dataset has bridged the gap between the SAD dataset and the actor identification dataset. We also proposed ASAD evaluation metrics by considering multi-label actions and actor identification. It is the first evaluation metric in ASAD such a complicated task. 

Besides the above success, it is important to note that our ASAD study also suffers some limitations: 
\begin{itemize}
\item Considering the high annotation cost, the size of our proposed ASAD dataset is still relatively small. Meanwhile, since the definition of action labels could be ambiguous, the action annotation may not be accurate. For instance, it is difficult to judge the boundary between ``walk'' and ''running'' in the continuous temporal domain. Or, without including the audio information, it is challenging to decide who is speaking. Such issues may impair the ASAD study. To cope with this issue, it is necessary to perform high-quality annotations with more annotators involved. 

\item Because evaluating the ASAD result is complicated, we separately evaluated spatial detection, actor identification, and multi-label action classification. Consequently, the overall ASAD performance is represented by multiple metric values. However, in an ideal case, we hope to utilize a single metric value to represent the overall ASAD performance. Considering that each of our ASAD metrics (\eg, HL@0.5) is obtained from a complex formula, it is challenging to integrate them into a single metric value. To find a solution, further exploration is needed. Since we have raised this question in the ASAD task, it might be solved in future works.
\end{itemize}
And we are considering addressing those issues in future works.

This paper is not the end, but rather the starting steps, we are excited to engage with the research community to explore ASAD deeper. We believe considering actor identification with spatiotemporal action detection could promote the research on video understanding and beyond. 



\printcredits

\bibliographystyle{cas-model2-names}
\bibliography{cas-refs}

\end{document}